\documentclass[letterpaper]{article} 
\usepackage{aaai2026}  
\usepackage{times}  
\usepackage{helvet}  
\usepackage{courier}  
\usepackage[hyphens]{url}  
\usepackage{graphicx} 
\usepackage{natbib}  
\usepackage{caption} 
\usepackage{algorithm}
\usepackage{algorithmic}

\usepackage{newfloat}
\usepackage{listings}
\DeclareCaptionStyle{ruled}{labelfont=normalfont,labelsep=colon,strut=off} 
\floatstyle{ruled}
\newfloat{listing}{tb}{lst}{}
\floatname{listing}{Listing}
\usepackage{bbding}
\usepackage{multirow, booktabs}
\usepackage{adjustbox}
\usepackage{changepage}
\usepackage{bm}
\usepackage{graphicx}
\usepackage{booktabs}
\usepackage{amsmath,amssymb,amsfonts}
\usepackage{cleveref}
\usepackage{tabularx}
\usepackage{colortbl}
\usepackage{xspace}
\usepackage{array}      
\usepackage{xcolor}
\usepackage{makecell}
\usepackage{subcaption}

\usepackage{tabulary}
\newcolumntype{I}{!{\vrule width 1pt}}

\newcolumntype{x}[1]{>{\centering\arraybackslash}p{#1pt}}
\newcolumntype{y}[1]{>{\raggedright\arraybackslash}p{#1pt}}
\newcolumntype{z}[1]{>{\raggedleft\arraybackslash}p{#1pt}}
\newlength\savewidth
\newcommand{\tablestyle}[2]{\setlength{\tabcolsep}{#1}\renewcommand{\arraystretch}{#2}\centering\footnotesize}

\definecolor{mygray}{RGB}{240,240,240}  
\usepackage{enumitem}
\usepackage{pifont}

\makeatletter
\newcommand{\thickhline}{%
	\noalign {\ifnum 0=`}\fi \hrule height 1pt
	\futurelet \reserved@a \@xhline
}
\makeatother

\newcommand{\ourmethod}{\textbf{\texttt{AbductiveMLLM}}\xspace}
\newcommand{\reasoner}{{\textsc{Reasoner}}\xspace}
\newcommand{\imaginer}{{\textsc{Imaginer}}\xspace}
\newcommand{\mllms}{{MLLMs}\xspace}
\newcommand{\chg}{{\textsc{CHG}}\xspace}

\newcommand{\ie}{\textit{i.e.}}
\newcommand{\eg}{\textit{e.g.}}
\newcommand{\etc}{\textit{etc.}}

\title{\textsc{AbductiveMLLM}: Boosting Visual Abductive Reasoning Within MLLMs}
\author{
   Boyu Chang\textsuperscript{\rm 1}, 
   Qi Wang\textsuperscript{\rm 1,2}, 
   Xi Guo\textsuperscript{\rm 1}, 
   Zhixiong Nan\textsuperscript{\rm 3}, 
   Yazhou Yao\textsuperscript{\rm 4}, 
   Tianfei Zhou\textsuperscript{\rm 1,5}\thanks{Corresponding author} 
}
\affiliations{
    \textsuperscript{\rm 1} Beijing Institute of Technology\\
    \textsuperscript{\rm 2} Beijing Institute of Technology, Zhuhai\\
    \textsuperscript{\rm 3} Chongqing University\\
    \textsuperscript{\rm 4} Nanjing University of Science and Technology\\
    \textsuperscript{\rm 5} State Key Laboratory of Environment Characteristics and Effects for Near-space\\

}

\usepackage{bibentry}

\begin{document}

\maketitle

\begin{abstract}
 Visual abductive reasoning (VAR) is a challenging task that requires AI systems to infer the most likely explanation for incomplete visual observations. While recent \mllms develop strong general-purpose multimodal reasoning capabilities, they fall short in abductive inference, as compared to human beings. To bridge this gap, we draw inspiration from the interplay between verbal and pictorial abduction in human cognition, and propose to strengthen abduction of \mllms by mimicking such dual-mode behavior.  Concretely, we introduce \ourmethod comprising of two synergistic components:  \reasoner and  \imaginer. The \reasoner operates in the verbal domain. It first explores a broad space of possible explanations using a blind LLM and then prunes visually incongruent hypotheses based on cross-modal causal alignment. The remaining hypotheses are introduced into the MLLM as targeted priors, steering its reasoning toward causally coherent explanations. The \imaginer, on the other hand, further guides \mllms by emulating human-like pictorial thinking. It conditions a text-to-image diffusion model on both the input video and the \reasoner’s output embeddings to “imagine” plausible visual scenes that correspond to verbal explanation, thereby enriching \mllms' contextual grounding. The two components are trained jointly in an end-to-end manner. Experiments on standard VAR benchmarks show that  \ourmethod achieves state-of-the-art performance, consistently outperforming traditional solutions and advanced MLLMs.
\end{abstract}

\begin{links}
    \link{Code}{https://github.com/ChangPtR/AbdMLLM}
\end{links}

\section{Introduction}\label{sec:intro}
Visual abductive reasoning is the process of forming an explanatory hypothesis for incomplete visual observations. 
It is an integral part of human cognition~\cite{peirce1935collected, shanahan2005perception} and  humans routinely employ it in everyday life, 
both \textit{verbally} and \textit{pictorially}~\cite{thagard1997abductive}. Given the observation $O$: \texttt{`the street is wet and the roof is dry'}, 
one might verbally abduce that a water truck has recently passed by and sprayed the street, based on some hidden governing rules 
such as \texttt{`rain wets both streets and roofs'} and \texttt{`a water truck wets only the street'}. 
Alternatively, one might pictorially abduce by forming a mental picture of a water truck  spraying water as it driving down the street. 
This imagined scene  resembles the hypothesized event in a  more direct and concrete way than a verbal or sentential representation would, 
and can in turn facilitates verbal abduction. 
This very ability gives humans a distinct advantage over machines in high-level reasoning, and represents one of the most valuable capacities to be emulated in modern machine vision system.

Recently, multimodal large language models (\mllms) have emerged as promising foundations for building  visual reasoning systems~\cite{wu2023multimodal}. Trained on vast amounts of human knowledge, these models have  developed impressive capabilities in multimodal reasoning tasks such as visual question answering, multimodal dialogue, and visual chart reasoning. However,  recent studies \cite{wang2024exploring,chinchure2024black} have highlighted a significant gap between current MLLMs and human abduction capability in understanding ambiguous observations.

To address this limitation, we draw inspiration from human cognition, where verbal and pictorial abduction interact to interpret incomplete visual observations. Based on this insight, we introduce \ourmethod, which enhances VAR capabilities of \mllms by integrating complementary verbal and pictorial abductive process. Specifically, our method consists of two synergistic components:  
1) a \reasoner, which extracts and selects high-quality verbal hypotheses from an LLM, serving as targeted priors for \mllms to generate plausible explanations;
2) a \imaginer, which simulates the pictorial thinking process using diffusion models to guide and refine MLLM-generated explanation.
These two components are jointly optimized in an end-to-end manner to allow interactions between verbal and pictorial modes of abductive thinking, and narrows the gap between abstract reasoning and concrete imagination in a more similar way as human cognition.

More specifically, \reasoner begins by prompting a LLM to generate a diverse set of candidate hypotheses based solely on video captions. While these hypotheses incorporate broad commonsense knowledge, they often lack grounding in the actual visual content and may therefore be causally inaccurate. To address this, we introduce a \textit{causality-aware contrastive learning} mechanism that promotes alignment between the observed video and causally relevant hypotheses (rather than relying on superficial similarity). This filtering process effectively prune out spurious candidates based on causal relevance and narrows the reasoning search space. The filtered hypotheses are then passed to an MLLM to generate a verbal explanation.
To complement verbal reasoning with visual imagination, \imaginer takes both the observed videos and output embeddings from the MLLM as conditioning signals for a generation model. 
Rather than training a new video generator from scratch, we adapt an existing text-to-image diffusion model (\ie, Stable Diffusion~\cite{rombach2022high}) with lightweight spatiotemporal adapters. As in~\cite{wang2024diffusion, ma2025genhancer}, the generator is not used to produce high-quality visual results, but instead serves as a reasoning guide: a latent denoising loss is applied to encourage the model to converge on visually plausible outcomes.

\textbf{Contributions.} This work presents \ourmethod, which represents a pioneering effort in enhancing the abductive capability of \mllms.
$\bullet$ From verbal perspective, we develop a causality-aware contrastive learning model to mine high-quality textual hypotheses, reducing reasoning space and providing crucial priors for \mllms. 
$\bullet$ From pictorial perspective, to the best of our knowledge, this is the first study to visual abductive reasoning that explicitly incorporates pictorial thinking capability to improve verbal abductions, inspired by human cognitive processes. 
Our method shows promising performance on standard benchmarks, consistently outperforming existing specialized small-scale models and  \mllms, setting the new state-of-the-art.

\section{Related Work}\label{sec:related_work}
\noindent\textbf{Visual Abductive Reasoning (VAR).}
VAR aims to infer the most likely explanation for partially observed visual events. Early VAR approaches addressed only static images. \cite{hessel2022abduction} introduced the Sherlock dataset and adapted CLIP for image-based abductive inference. RCA~\cite{zhang2024rca} augmented this by a visually guided multi-head attention mechanism and a revised contrastive loss. BiGED~\cite{tan2025inferring} proposed a relational GNN to predict human pre-action sequences in indoor scenes from a single image. However, these approaches are constrained by the static and incomplete nature of single-frame observations, and often fail to capture the complex spatiotemporal causal structure of open-world scenarios. 

To address the limitations above, recent works have shifted toward video-based abductive reasoning. REASONER~\cite{liang2022visual} is among the first to build a dataset of real-world visual event sequences, and combines a causality-aware video encoder with a cascade decoder, enabling abductive reasoning over arbitrary visual events. Subsequently, UPD-Trans~\cite{xu2024probabilistic} introduced probabilistic distillation in a Transformer. Conan~\cite{xu2023active} builds an agent to explore active interaction in simulated environments. Videoabc~\cite{zhao2022videoabc} applies hierarchical reasoning to capture long-term dependencies. Knowledge integration has also been explored: KN-VLM~\cite{tan2025kn} introduces visual knowledge from observed videos and textual knowledge from an external knowledge base. MAR~\cite{li2023multi} and COIN~\cite{li2024cross} incorporate symbolic reasoning to enhance abduction of human actions. 
All these studies are specialized small-scale models and focus exclusively on verbal reasoning. In light of recent advances in \mllms, we pivot to enhance \mllms for visual abductive reasoning. Drawing on human cognitive processes, we propose a unified multimodal network for video-based abductive reasoning, which integrates both verbal and pictorial thinking.

\begin{figure*}[ht]
	\centering
	\includegraphics[width=.99\textwidth]{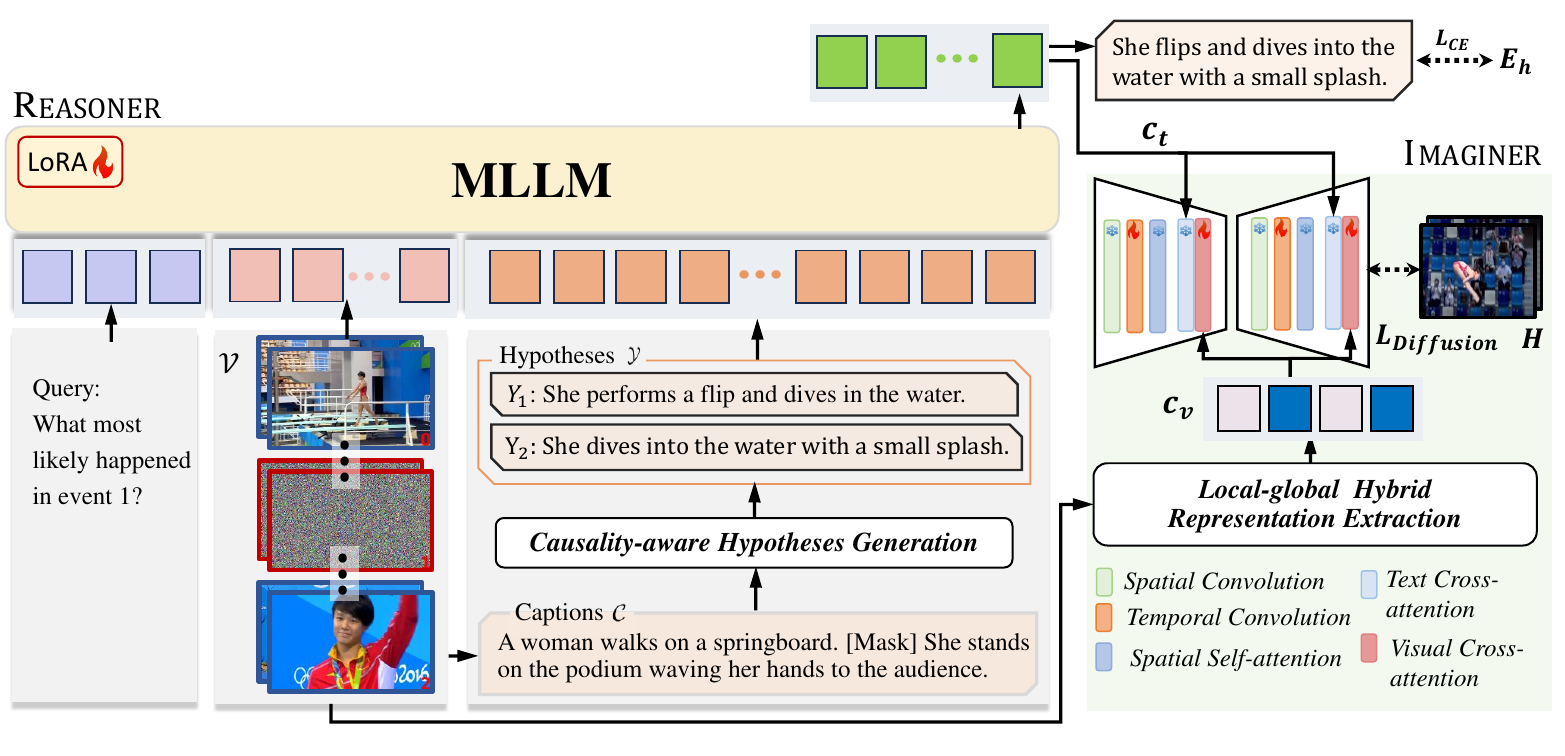}
	\caption{\textbf{Network architecture of \ourmethod.} The network consists two synergistic components: \reasoner and \imaginer. On the left, \reasoner takes a query and a sequence of incomplete observations $\mathcal{V}$ as input. First, it generates captions $\mathcal{C}$ for each observed event. Based on $\mathcal{C}$, Causality-aware Hypotheses Generation module provides high-quality hypotheses $\mathcal{Y}$ for the MLLM. Its output embeddings are passed to \imaginer as textual conditions $\bm{c}_t$ for imagination, which are also used to generate verbal abduction results. On the right, \imaginer is adapted from a text-to-image diffusion model through the integration of lightweight adapters. It takes $\bm{c}_t$ and $\bm{c}_v$ as multimodal conditions, where $\bm{c}_v$ is visual local-global hybrid representations extracted from the observations. The two components are trained end-to-end with $\mathcal{L}_\text{CE}$ and $\mathcal{L}_\text{Diffusion}$. } 
	\label{fig:pipeline}
\end{figure*}

\noindent\textbf{Multimodal Large Language Models (\mllms).} \mllms have emerged as leading paradigm for video understanding. Mainstream approaches typically build upon pre-trained large language models, integrate dedicated video encoders, and employ techniques such as self-supervised learning and instruction tuning to achieve effective vision-language alignment and enhance multimodal representation capabilities \cite{wang2024internvideo2,lin2024video,lyu2023macaw,chen2023x,zohar2024apollo,wang2024qwen2}. These models have been successfully applied to a wide range of multimodal tasks, including video question answering~\cite{Wang2024VideoCoT,Maaz2024VideoChatGPT,wang2025videorft}, multimodal dialogue \cite{Luo2023ValleyVA}, video captioning \cite{damonlpsg2024videollama2,Xu2023YoukumPLUGA1}, \etc. 
Nevertheless, recent studies \cite{wang2024exploring,chinchure2024black} have underscored a significant gap between current \mllms and human abductive capability, which indicates that these models still lack the capacity for advanced reasoning grounded in causal relationships.
Our work proposes enhancements from verbal and pictorial modalities, and narrows the abductive reasoning capability between \mllms and humans.

\section{Our Approach}\label{sec:method}
\noindent\textbf{VAR Task.} We follow the  task definition of VAR in~\cite{liang2022visual}. Given a video sequence containing $T$ events $\mathcal{V} \!=\! \{O_1, \dots, O_{t-1}, H, O_{t}, \dots, O_{T-1}\}$, where the events are logically related and chronologically organized. Among them, $\mathcal{O} = \{O_{t}\}_{t=1}^{T-1}$ denotes the collection of $T\!-\!1$ observed premise events, and $H$ represents  unobserved explanatory event. Notably, $H$ may occur at any position within $\mathcal{V}$. The goal of the VAR task is to infer the most likely verbal explanation $E_h$ for the unobserved event $H$, based on the observed events in $\mathcal{O}$.

\noindent\textbf{Main Idea.} Inspired by how humans integrate verbal and pictorial thinking for abductive reasoning, we introduce a joint network to enhance the abductive reasoning capabilities of \mllms. 
\ourmethod consists of two main modules: \reasoner and \imaginer.
\reasoner first generates candidate hypotheses with a blind LLM, then selects causally relevant hypotheses through cross-modal causal contrastive learning, which enhances \mllms' reasoning in verbal mode.
\imaginer is a diffusion model with lightweight adapters, conditioning on \reasoner output embeddings and observations. It is trained end-to-end with \reasoner to provide enhancement in pictorial mode. 
Fig.~\ref{fig:pipeline} illustrates the entire process of our method.

\subsection{\reasoner : Abduction in Verbal Mode}
\label{sec1}
The \reasoner enhances the abduction of \mllms in verbal mode. It first generates candidate hypotheses using a blind LLM. Then, a causality-aware hypotheses selection module is proposed to  prune visually irrelevant hypotheses based on causal relevance. The remaining hypotheses are subsequently passed to \mllms as targeted priors for verbal abduction.  

\subsubsection{Causality-aware Hypotheses Generation (\chg)}\mbox{}\\[0pt]
\label{sub:gen}
\textit{Step 1: Candidate Verbal Hypotheses Generation.}
Abductive reasoning presents a significant challenge due to its vast and complex space of plausible explanations. To alleviate this, we leverage the knowledge-rich capabilities of advanced LLMs to generate a diverse set of candidate hypotheses, thereby narrowing the reasoning space. Specifically, we first employ a pre-trained MLLM to generate video captions for each observed video clip in $\mathcal{O}$, yielding a sequence of  descriptions $\mathcal{C}\!=\!\{C_{t}\}_{t=1}^{T-1}$, with each $C_{t}$ corresponds to the description of $O_{t}$. Afterwards, we prompt GPT-4o-mini~\cite{hurst2024gpt} to infer plausible missing events under the instruction:  \textit{You are an event-completion expert, infer the most plausible event at the [MASK] position.}. To ensure diversity and reduce redundancy, we query GPT-4o-mini multiple times per video instance using a relatively high sampling temperature (\eg, 1.4). This finally results in a collection of $L$ diverse candidate hypotheses denoted as $\mathcal{Y}=\{Y_{i}\}_{i=1}^{L}$, where each $Y_i$ represents a distinct verbal explanation for the missing event.

\textit{Step 2: Causality-aware Cross-modal Hypotheses Selection.}
Since the hypotheses in $\mathcal{Y}$ are derived solely from textual captions, they may include low-quality, hallucinated candidates that hinder effective reasoning in MLLMs. Therefore, we introduce a contrastive learning based hypotheses selection module to identify causally relevant hypotheses from $\mathcal{Y}$ by leveraging $\mathcal{O}$. Unlike standard contrastive learning, which merely establishes superficial similarity between visual and textual modalities, this module is specifically designed to capture the causal relevance between visual observations and textual hypotheses.

Sufficient high-quality negative samples are crucial in contrastive learning~\cite{chen2020simple, robin2021contrast}. While the ground-truth explanation $E_h$ could directly serve as the positive hypothesis, constructing diverse and semantically meaningful negative hypotheses remains a challenge. To generate these negative hypotheses, we utilize GPT-4o-mini~\cite{hurst2024gpt}. The prompt begins with a task description:
\textit{There is a contrastive learning task aimed at matching the missing video caption using a series of observed videos.} We then provide GPT with the positive explanation $E_h$, the observed captions $\mathcal{C}$, and instructions: \textit{The negative samples should differ in semantics from the positive sample, but still align with the logical context of observed captions.} 
We call GPT-4o-mini multiple times to obtain $M$ negative hypotheses.

\begin{figure}[t]
	\centering
	\includegraphics[width=0.48\textwidth]{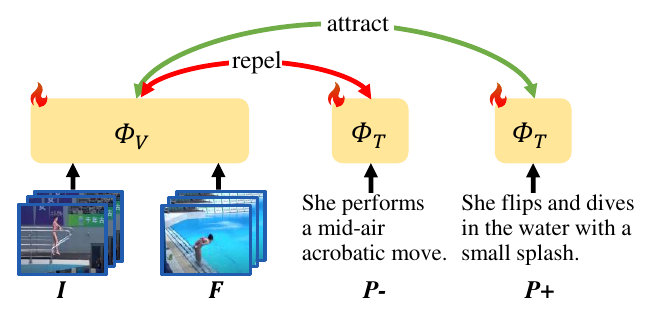}
	\caption{\textbf{Illustration of cross-modal causal contrastive learning.} Encoders $\Phi_V$, $\Phi_T$ learn to attract causally plausible hypotheses $\mathcal{P}^+$ to the observations, and repel causally irrelevant hypotheses $\mathcal{P}^-$.}
	\label{fig:causal}
\end{figure}

Given any input $\mathcal{V}$, it can be naturally partitioned into three sequential segments: the \textit{initial} segment $\mathcal{I}$, the \textit{process} segment $\mathcal{P}$, and the \textit{final} segment $\mathcal{F}$, and $H$ may correspond to any of them. In the following, we take the case where $H$ serves as the \textit{process} segment as an illustrative example. The observations before and after $H$ are then treated as the \textit{initial} and \textit{final} segments. 
As illustrated in Fig.~\ref{fig:causal}, the module comprises a vision encoder $\Phi_V$ and a text encoder $\Phi_T$, which project embeddings of these segments into a joint causal space. Specifically, $\Phi_V$ encodes the observed \textit{initial} and \textit{final} segments into visual embeddings $X_\mathcal{I}$ and $X_\mathcal{F}$. $\Phi_T$ encodes the postive and negative hypotheses into textual embedding $X_\mathcal{P}^+$ and $X_\mathcal{P}^-$.

During training, the model is optimized with a contrastive objective that maximizes the causal relevance between the observed video and the positive hypothesis, while minimizing the relevance with negative hypotheses. This is achieved via the NT-Xent loss~\cite{chen2020simple}:
\begin{equation}\small
	\label{loss_nt}
	\mathcal{L}_\text{Contrast} = -\log \frac{\exp\left(\langle \bm{X}_\mathcal{I} + \bm{X}_\mathcal{P}^{+}, \bm{X}_\mathcal{F} \rangle / \tau \right)}
	{\sum_{i=1}^{M} \exp\left( \langle \bm{X}_\mathcal{I} + \bm{X}_\mathcal{P}^{i-,+}, \bm{X}_\mathcal{F}\rangle / \tau \right)},
\end{equation} 
where \( \tau \) is the temperature coefficient, $X_\mathcal{P}^{i-}$ is the embedding of the $i$-th negative hypothesis, and \( \langle \cdot, \cdot \rangle \) is cosine similarity between embeddings.

During inference, for each candidate hypothesis $Y_i \in \mathcal{Y}$ in Step 1, we project it into the joint space with $\Phi_T$ and compute its causal relevance score to the observed videos:
\begin{equation}\small
\text{Score}(Y_i) =\langle \bm{X}_\mathcal{I} + \bm{X}_{Y_i}, \bm{X}_\mathcal{F} \rangle,
\end{equation} 
where $\bm{X}_{Y_i} = \Phi_T(Y_i)$ denotes the embedding of $Y_i$. We then rank all candidate hypotheses based on their scores and select the top-$k$ most causally aligned hypotheses for downstream reasoning.

\subsubsection{Hypotheses Guided Verbal Abduction in \mllms}\mbox{}\\[0pt]
To supply the MLLM with high-quality hypotheses, we retain only top‑$k$ highest‑scoring hypotheses in the prompt for MLLM reasoning.
We concatenate multiple events within $\mathcal{V}$ into a single video.  
For the unobserved event \( H \), we fill the gap with placeholder frames with random pixels.  
To explicitly inform the MLLM of the position of \( H \), we add numbers on video frames tailored for our VAR task.  
Instead of frame-level indices in Number-Prompt~\cite{wu2024number}, these numbers encode event-level indices, 
allowing the model to understand the temporal ordering between different events.  
The output embeddings of \reasoner (denoted by $\bm{c}_t$) is subsequently passed to \imaginer as conditions for imagination. $\bm{c}_t$ is further decoded into verbal explanation as abduction result of \ourmethod.

\subsection{\imaginer : Abduction in Pictorial Mode}
\label{sec3}

\begin{figure}[t]
	\centering
	\includegraphics[width=0.49\textwidth]{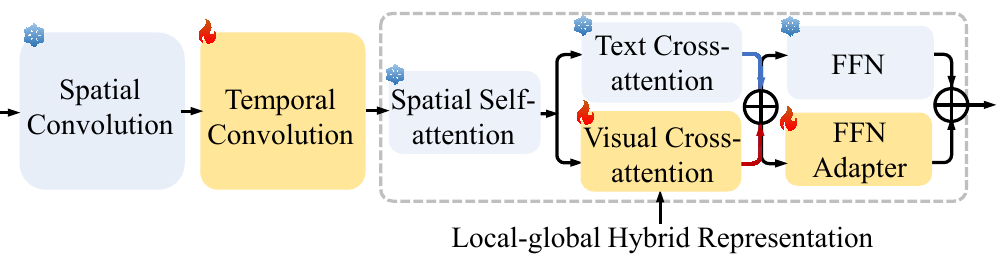}
	\caption{\textbf{A standard U-Net block of Stable Diffusion with proposed adapters.} During training, we only update parameters of the adapters (in yellow), and freeze parameters of other modules (in blue).}
	\label{fig:gen}
\end{figure}

The \imaginer serves to provide richer contextual cues to facilitate the abduction of \mllms in pictorial mode. Notably, as our objective centers on extracting visual guidance rather than optimizing video generation, we implement \imaginer by extending a text-to-image diffusion model (\ie, Stable Diffusion) to video generation through lightweight adapters. As shown in Fig.~\ref{fig:gen}, we introduce three kinds of adapters to each U-Net block in Stable Diffusion.

\textbf{Visual Cross-attention Adapter (V-Adapter).} In vanilla Stable Diffusion, the U-Net attention blocks perform only self-attention for individual frames, neglecting the information from other frames. In our case, it fails to leverage the observed visual cues in $\mathcal{O}$, which potentially contain valuable information relevant to textual explanations. To address this issue, we introduce V-Adapter to inject informative visual priors into the model.

Directly incorporating all frames from $\mathcal{O}$, however, is computationally expensive and prone to noise due to redundancy. To overcome this, we propose an efficient strategy to extract a  \textit{local-global hybrid representation} from $\mathcal{O}$ that captures both  fine-grained and holistic visual semantics relevant to the textual explanation $E_h$. \textit{For the local representation}, we  employ CLIP's image and text encoders to obtain embeddings $\{\bm{c}_v^i\}_{i=1}^{N}$ for all $N$ frames in $\mathcal{O}$, and $\bm{c}_h$ for $E_h$. Then we calculate the similarity scores:
\begin{equation}\small
    \gamma^i = \frac{\exp\left(\operatorname{sim}(\bm{c}_v^i, \bm{c}_h)\right)}{\sum_{j=1}^N \exp\left(\operatorname{sim}(\bm{c}_v^j, \bm{c}_h)\right)} \in [0,1],
\end{equation}
and only concatenate high-scoring frames together to form the local representation  $\bm{c}_{local}$.
\textit{For the global representation}, we compute a weighted average of $\{\bm{c}_v^i\}_{i=1}^{N}$ based on the similarity scores $\{\gamma^i\}_{i=1}^{N}$, yielding the  representation $\bm{c}_{global}=\sum_{i=1}^N\gamma^i\bm{c}_v^i$. 
Finally, we concatenate the local and global representations as the visual condition, denoted as $\bm{c}_v = [\bm{c}_{local}; \bm{c}_{global}]$. Hence, V-Adapter can be formulated as:
\begin{equation}\small
\text{V-Adapter}(\bm{Q},\bm{K}_v,\bm{V}_v)=
\operatorname{Softmax}\!\Bigl(\frac{\bm{Q}\,\bm{K}_v^{\top}}{\sqrt{d_k}}\Bigr)\,\bm{V}_v,
\end{equation}
where $\bm{Q}\!=\!\bm{x}\bm{W}^q,\bm{K}_v \!=\! \bm{c}_v \bm{W}_v^k, \bm{V}_v \!=\! \bm{c}_v \bm{W}_v^v$. 
$\bm{K}_v$/$\bm{V}_v$ are the key/value in cross-attention computation of V-Adapter, $\bm{W}^q$/$\bm{W}_{v}^{k,v}$ are projection matrices.

We integrate the V-Adapter in parallel with the original text cross-attention in each U-Net block, while freezing the textual attention parameters.  This enables the model to attend to both visual and textual cues simultaneously.  
The outputs from the parallel cross-attention branches are then summed. The process is depicted as follows:
\begin{equation}\small
\bm{x}
= \text{CrossAttn}(\bm{Q},\bm{K}_t,\bm{V}_t)
 + \text{V-Adapter}(\bm{Q},\bm{K}_v,\bm{V}_v),
\end{equation}
where $\bm{K}_t \!=\! \bm{c}_t \bm{W}_t^k, \bm{V}_t \!=\! \bm{c}_t \bm{W}_t^v$, 
$\bm{K}_t$/$\bm{V}_t$ are the key/value of the textual cross-attention, $\bm{W}_{t}^{k,v}$ are projection matrices.

\textbf{Temporal Convolution Adapter (T-Adapter).} The T-Adapter is designed to model temporal dependencies across frames and is appended after the spatial convolution layers in each U-Net block. It uses depth-wise 3D convolutional in a projected lower-dimensional space, which can alleviate the complexity of temporal modeling~\cite{singer2022make,blattmann2023align,xing2024simda}. 
To keep structural consistency and further improve temporal modeling, we adopt a fully convolutional design, which is defined as:
\begin{equation}\small
    \text{T-Adapter}(\bm{x}) = \bm{x} + \text{Conv3D}_{up}\left(\text{Conv3D}_{down}(\bm{x}) \right),
\end{equation}
where $\text{Conv3D}_{up}$ and $\text{Conv3D}_{down}$ denote the up-projection and down-projection layers, both are 3D convolutions.

\textbf{FFN Adapter (F-Adapter).} The F-Adapter enhances spatial representation while preserving the integrity of the original feed-forward network. It is introduced as a parallel block to the FFN layer, making sure that the pretrained FFN remains unchanged while adapting to the spatial features of videos.
FFN adapter consists of two fully connected (FC) layers with GELU activation~\cite{xing2024simda}, which can be formulated as:
\begin{equation}\small
    \text{F-Adapter}(\bm{x}) = \bm{x} + \text{FC}_{up}(\text{GELU}(\text{FC}_{down}(\bm{x}))),
\end{equation}
where $\text{FC}_{up}$ and $\text{FC}_{down}$ are the up-projection and down-projection layers.

\subsection{Network Training}
\label{sec:training}
We adopt a two-stage training paradigm in which the modules are first trained independently before undergoing joint end-to-end optimization. In Stage I, for \reasoner, the MLLM is finetuned with LoRA under the standard cross-entropy loss $\mathcal{L}_{\textit{CE}}$, enabling it to generate plausible hypotheses from incomplete observations. For \imaginer, we freeze the weights of stable diffusion and only update the parameters of adapters using the same conditional latent diffusion loss $\mathcal{L}_{\textit{Diffusion}}$ in \cite{rombach2022high}.
We also apply Min-SNR weighting strategy~\cite{hang2023efficient}, 
which adaptively reweights the loss at each diffusion timestep to accelerate the convergence of the diffusion model. In Stage II, \reasoner and \imaginer are jointly tuned in an end-to-end manner. The overall loss is defined as:
\begin{equation}\small
\label{eq:e2e_loss}
    \mathcal{L} = \mathcal{L}_{\text{CE}} + \alpha  \mathcal{L}_{\text{Diffusion}},
\end{equation} 
where $\alpha$ is a  coefficient that balances the two terms.

\begin{table*}[t]
	\centering
	\tablestyle{16pt}{1.1}
	\begin{tabular}{l||ccccc}
		\thickhline\rowcolor{mygray}
		Method   & BLEU@4 & METEOR & ROUGE & CIDEr  & BERT-S \\
		\hline
		Human     & 11.35  & 19.36  & 36.92 & 147.79 & 40.59  \\ \hline
		$\bullet$ \textbf{Traditional Models} \\
		VTrans~\cite{zhou2018end}     & 0.71   & 6.92   & 19.12 & 7.11   & 22.13  \\
		MFT~\cite{xiong2018move}        & 1.81   & 7.16   & 19.16 & 17.67  & 25.90  \\
		Trans-XL~\cite{dai2019transformer}   & 2.96   & 7.51   & 20.94 & 24.54  & 27.23  \\
		MART~\cite{lei2020mart}       & 2.86   & 7.47   & 20.87 & 24.05  & 27.77  \\
		PDVC~\cite{wang2021end}       & 3.00   & 8.54   & 20.71 & 25.14  & 27.80  \\
		REASONER~\cite{liang2022visual}  & 3.44   & 9.05   & 22.89 & 30.75  & 30.64  \\
		KN-VLM~\cite{tan2025kn}    & 4.72   & 10.74  & 24.40 & 37.20  & 33.17  \\
		UPD-Trans~\cite{xu2024probabilistic} & 5.40   & 11.16  & 25.62 & 41.66  & 30.80  \\ \hline
		$\bullet$ \textbf{\mllms} \\
		GPT-4o-mini~\cite{hurst2024gpt} & 0.63   & 7.38  & 13.64 & 7.30  & 12.27  \\
		VideoChat2-7B~\cite{li2024mvbench} & 1.24   & 7.55  & 17.06 & 19.51  & 26.40  \\
		Qwen2VL-7B~\cite{wang2024qwen2} & 2.41   & 11.29  & 21.61 & 29.25  & 30.01  \\
		Qwen2VL-7B$^\texttt{FT}$~\cite{wang2024qwen2} & 5.67   & 12.77  & 27.11 & 50.82  & 36.03  \\
		\textbf{\ourmethod} & \textbf{6.54} & \textbf{13.41} & \textbf{27.95} & \textbf{57.04} & \textbf{36.80} \\
		\hline
	\end{tabular}
	\caption{Quantitative results on VAR \texttt{test}. \texttt{FT} means the model is finetuned on the dataset.}
	\label{tab:var_scores}
\end{table*}

\begin{table*}[t]
	\centering
	\tablestyle{16pt}{1.1}
	\begin{tabular}{l||ccccc}
		\thickhline\rowcolor{mygray}
		Method   & BLEU@4 & METEOR & ROUGE & CIDEr  & BERT-S \\
		\hline
		$\bullet$ \textbf{Traditional Models} \\
		REASONER~\cite{liang2022visual}  & 3.54   & 9.47   & 24.62 & 32.99  & 23.19  \\ \hline
		$\bullet$ \textbf{\mllms} \\
		GPT-4o-mini~\cite{hurst2024gpt} & 0.45   & 4.22  & 12.78 & 14.15  & 9.75  \\
		VideoChat2-7B~\cite{li2024mvbench} & 0.49   & 4.59  & 14.31 & 17.82  & 7.96  \\
		Qwen2VL-7B~\cite{wang2024qwen2} & 2.46   & 8.41  & 22.10 & 35.83  & 21.83  \\
		Qwen2VL-7B$^\texttt{FT}$~\cite{wang2024qwen2} & 5.66   & 12.62  & 28.64 & 68.44  & 29.09  \\
		\textbf{\ourmethod} & \textbf{6.16} & \textbf{13.46} & \textbf{30.06} & \textbf{77.70} & \textbf{30.77} \\
		\hline
	\end{tabular}
	\caption{Quantitative results on YouCookII \texttt{test}. \texttt{FT} means the model is finetuned on the dataset.}
	\label{tab:cook_scores}
\end{table*}

\section{Experiment}\label{sec:experiment}
\subsection{Experimental Setup}
\textbf{Network Architecture.} 
In \reasoner, the MLLM is implemented as Qwen2VL-7B-Instruct~\cite{wang2024qwen2}. 
$\Phi_V$ consists of pretrained ResNet200~\cite{he2016deep}/BN-Inception~\cite{ioffe2015batch} as in ~\cite{liang2022visual} and trainable 2-layer Transformer encoder. $\Phi_T$ consists of pretrained CLIP text encoder and trainable 2-layer MLP.
In \imaginer, Stable Diffusion-v1-4~\cite{rombach2022high} is the backbone, with $256\times256$ resolution and $32\times32$ latent size. 
The above two modules is connected by the Bridge Layer, which is implemented as a 2-layer MLP with SiLU activation.

\noindent\textbf{Dataset.} We validate our approach on two datasets:

\begin{itemize}
\item \textbf{VAR} \cite{liang2022visual}. 
The VAR dataset contains 8,606 annotated samples sourced from 3,718 unique videos. 
Each video includes an average of 4.17 events, each lasting approximately 37.8 seconds. 
The dataset is split into train/val/test splits, containing 7,053/460/1,093 samples. 

\item \textbf{YouCookII} \cite{zhou2018towards}. YouCookII is a large-scale cooking video dataset containing 1,333/457/210 videos for train/val/test. We follow the setup of \cite{tan2025kn} to organize the videos for the VAR task. Specifically, we re-partition the original training and validation sets to obtain 1,533/257 videos for train/test. For each video, we iteratively select one event as the explanation event and treat the remaining events as observed events. Finally, we obtain 11,737/1,870 data samples for train/test.

\end{itemize}

\noindent\textbf{Competitor.}
We present a comprehensive comparison between \ourmethod and  state-of-the-art models, including traditional specialized small-scale models~\cite{liang2022visual,tan2025kn,xu2024probabilistic}, proprietary and open-source MLLMs~\cite{hurst2024gpt,li2024mvbench,wang2024qwen2}. Each traditional model is trained separately on \texttt{train} sets of VAR and YouCookII datasets. We also finetune Qwen2VL-7B-Instruct on the datasets for the same total epochs as our baseline (Qwen2VL-7B$^\texttt{FT}$).

\noindent\textbf{Training Configuration.}
In Stage I, we first finetune the MLLM and the diffusion model separately for 2 epochs. To train the cross-modal constrastive learning module, we generate 100 hard negatives for each positive $E_h$. Then we train it for 10 epochs and use the model with the highest training accuracy to prune candidate hypotheses. We set $k=3$ in the top-$k$ hypotheses selection.
In Stage II, we finetune the \reasoner and \imaginer in an end-to-end manner for 1 epoch. All models are trained on 4 A800 GPUs with 80GB memory per-card.

\noindent\textbf{Metric.} We follow prior works~\cite{liang2022visual,tan2025kn,xu2024probabilistic} to use five well-known automated metrics \ie, BLEU@4, METEOR, ROUGE-L, CIDEr and BERT-S~\cite{zhang2019bertscore} for evaluation. 

\begin{figure*}[t]
	\centering
	\includegraphics[width=.99\textwidth]{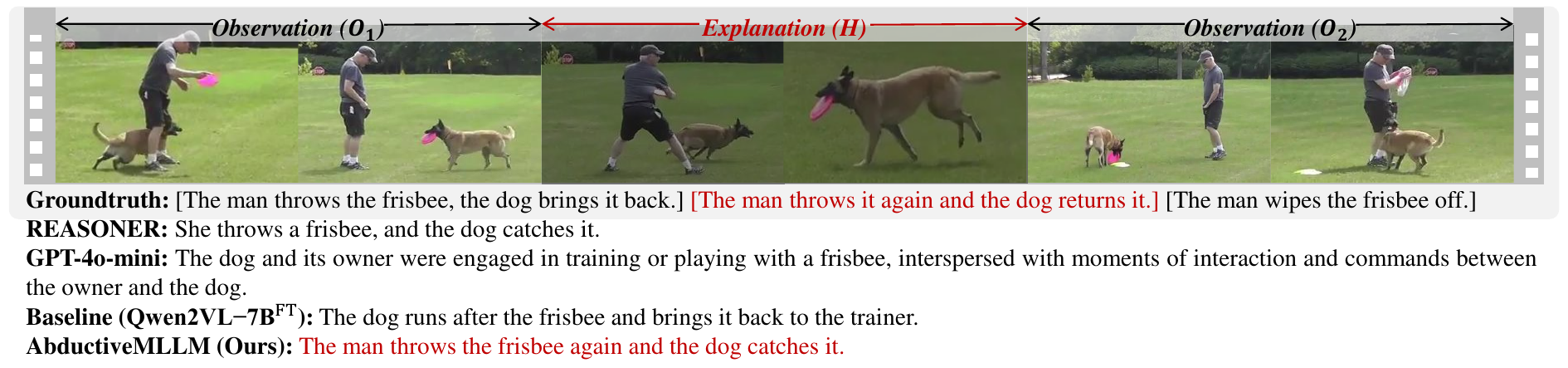}
	\caption{Qualitative comparison of \ourmethod on an example from VAR \texttt{test}.}
	\label{fig:res}
\end{figure*}

\subsection{Main Result}
\label{performance}
\noindent\subsubsection{Quantitative Result.} As shown in Table~\ref{tab:var_scores} and Table~\ref{tab:cook_scores}, \ourmethod achieves the best performance across all metrics on both VAR \texttt{test} and YouCookII \texttt{test} benchmarks, demonstrating consistent advantages over both traditional models and advanced MLLMs.

Several key observations can be drawn from the results. First, compared to the best traditional model (\ie, UPD-Trans), our method achieves significantly higher scores across all metrics on VAR, including \textbf{+1.14} BLEU@4, \textbf{+2.25} METEOR, \textbf{+2.33} ROUGE, \textbf{+15.38} CIDEr, \textbf{+6.00} BERT-S.
This demonstrates that MLLMs, when equipped with abductive reasoning module, can outperform task-specific architectures, offering a more scalable solution for VAR. Second, our model also surpasses all zero-shot MLLMs (\ie, GPT-4o-mini, VideoChat2-7B, Qwen2VL-7B), with improvements of over \textbf{+6.34} ROUGE, \textbf{+27.79} CIDEr,  \textbf{+6.79} BERT-S on VAR, and \textbf{+7.96} ROUGE, \textbf{+41.87} CIDEr, \textbf{+8.94} BERT-S on YouCookII. These results reveals the lack of abductive reasoning capabilities in existing general LLMs.
Even when compared to an MLLM specifically fine-tuned for the VAR task (Qwen2VL-7B$^\texttt{FT}$), our method achieves consistent gains on both datasets (\eg, \textbf{+0.84} ROUGE, \textbf{+6.22} CIDEr on VAR, and \textbf{+1.42} ROUGE, \textbf{+9.26} CIDEr on YouCookII), confirming that the integration of verbal and pictorial abduction provides complementary VAR capability beyond tuning alone can offer. 
Third, we observe that there still remains a significant gap between human abduction and AI models, which indicates that AI models still have substantial room to grow before reaching human-level cognitive capabilities.

\noindent\subsubsection{Qualitative Result.}
Fig.~\ref{fig:res} contains the explanatory hypotheses from \ourmethod and other competitors~\cite{liang2022visual,hurst2024gpt,wang2024qwen2} as well as groundtruth sentences. 
We can find that our method is able to discover and correctly describe the cause-effect chain, and hence generate a plausible hypothesis: \texttt{`throws the frisbee again and the dog catches it'}, that well explans the observed events. In contrast, other competitors typically produce unsatisfactory results, \eg, REASONER~\cite{liang2022visual} misidentifies the person's gender, GPT-4o-mini offers only \textit{general descriptions} rather than \textit{fine-grained reasoning}, baseline (Qwen2VL-7B$^\texttt{FT}$) fails to infer the man's action.

\subsection{Ablation Study}
\label{ablation}
We conduct ablative experiments on VAR \texttt{test} for in-depth analyzing each design in our approach. 

\subsubsection{Key Component Analysis.} We first study the efficacy of core model designs in Table~\ref{tab:e2e}. The first row gives the performance of the baseline (Qwen2VL-7B$^\texttt{FT}$).
The results in the second and third rows reveal that both \chg and \imaginer can improve all five metrics, and the larger gains are achieved on overlap-sensitive scores (BLEU@4, CIDEr), indicating that both designs can help the model produce essential content words.
Furthermore, by comparing the second and third rows, \imaginer can bring more gains on semantics-oriented metrics (METEOR, ROUGE) and the embedding-based BERT-S, suggesting richer, visually grounded verbal results. 
From the last row, we can conclude that combining the two designs together leads to the best results.

\begin{table}[t]
\centering
\tablestyle{1pt}{1.1}
\begin{tabular}{c c||c c c c c c}
  \thickhline 
  \rowcolor{mygray}
  \makecell{\chg} & \makecell{\imaginer}
    & BLEU@4 & METEOR & ROUGE & CIDEr & BERT-S       \\ 
  \hline
                   &                     &  5.67  & 12.77  & 27.11 & 50.82 & 36.03   \\
  \ding{51}        &                     &  6.33  & 12.96  & 27.21 & 53.60 & 36.31   \\
                    & \ding{51}           &  6.35  & 13.07  & 27.52 & 55.00 & 36.40   \\
  \ding{51}        & \ding{51}           & \textbf{6.54} & \textbf{13.41} & \textbf{27.95} & \textbf{57.04} & \textbf{36.80} \\
  \hline
\end{tabular}
\caption{Diagnostic experiments for \ourmethod.}
\label{tab:e2e}
\end{table}

\subsubsection{Top-$k$ Hypotheses Selection.} 
As we select the top-$k$ candidate hypotheses from GPT, we study the impact of $k$ in the hypotheses selection.  
As shown in Table~\ref{tab:select}, $k = 0$ means training without any hypothesis from GPT. We observe that when $k$ is larger than 3, excessive candidate hypotheses leads to a noticeable drop in performance across all metrics; when $k$ is less than 3, insufficient hypotheses fails to provide noticeable gains in performance. We therefore use $k=3$ for all experiments.

\begin{table}[t]
\centering
\tablestyle{5pt}{1.1}
\begin{tabular}{c||ccccc}
    \thickhline
    \rowcolor{mygray}
$k$   & BLEU@4 & METEOR & ROUGE & CIDEr & BERT-S \\ \hline
0 & 6.35 & 13.07   & 27.52  & 55.00 & 36.40 \\
1 & 6.41   & 12.94  & 27.52 & 54.15 & 36.31  \\
\textbf{3}  & \textbf{6.54} & \textbf{13.41} & \textbf{27.95} & \textbf{57.04} & \textbf{36.80}  \\
6 & 6.32 & 13.13 & 27.68  & 54.89  & 36.47          \\
10 & 6.29 & 13.22 & 27.66  & 53.66  & 36.40            \\
 \bottomrule
\end{tabular}
\caption{Ablation study on top-$k$ hypotheses selection.}
\label{tab:select}
\end{table}

\subsubsection{Coefficient $\alpha$.} 
We study the impact of $\alpha$ in Eq.~\ref{eq:e2e_loss} in Table~\ref{tab:alpha}.
A larger $\alpha$ indicates a greater degree of intervention by imagination in the training.
Among the various $\alpha$ values we examined, the best performance is reached at $\alpha = 5$.
Nonetheless, \ourmethod maintains relatively stable performance with different $\alpha$ values, indicating that our model's performance is not sensitive to the selection of $\alpha$.

\begin{table}[t]
    \centering
    \tablestyle{4pt}{1.1}
    \begin{tabular}{c||ccccc}
        \thickhline
        \rowcolor{mygray}
        $\alpha$ (Eq.~\ref{eq:e2e_loss})   & BLEU@4 & METEOR & ROUGE & CIDEr & BERT-S \\ \hline
        1 & 6.33 &  13.36  & 27.86  & 54.91 & 36.76 \\
        3 & 6.52 &  13.39  & \textbf{28.11}  & 56.54 & \textbf{36.86} \\
    \textbf{5}  & \textbf{6.54} & \textbf{13.41} & 27.95 & \textbf{57.04} & 36.80  \\
        7 & 6.42 & 13.40 & 27.90  & 55.32  & 36.75            \\
        9 & 6.50 & 13.32 & 27.86  & 55.52  & 36.65           \\
     \bottomrule
    \end{tabular}
    \caption{Ablation study on $\alpha$.}
    \label{tab:alpha}
\end{table} 

\subsubsection{Adapters in \imaginer.} 
We assess the contribution of the proposed adapters in \imaginer. 
As shown in Table~\ref{tab:adapter}, individually removing each adapter from \imaginer leads to a performance drop, especially on the CIDEr metric, which indicates the model produces fewer accurate key terms in verbal abduction. The results confirm the necessity of each adapter in \imaginer.
Moreover, training with variants of \imaginer still outperforms the model without \imaginer (the second row in Table~\ref{tab:e2e}), which further proofs the importance of pictorial thinking.

\begin{table}[t]
    \centering
    \tablestyle{1pt}{1.1}
    \begin{tabular}{c||ccccc}
        \thickhline
        \rowcolor{mygray}
        Variant   & BLEU@4 & METEOR & ROUGE & CIDEr & BERT-S \\ \hline
        \textbf{\ourmethod} & \textbf{6.54} & \textbf{13.41} & \textbf{27.95} & \textbf{57.04} & \textbf{36.80} \\
        w/o V-Adapter  & 6.36 &  13.29 & 27.76  & 54.51 & 36.68 \\ 
        w/o T-Adapter  & 6.42 &  13.28  & 27.74  & 54.99 & 36.68 \\ 
        w/o F-Adapter  & 6.47 &  13.31  & 27.70  & 54.52 & 36.63 \\ 
     \bottomrule
    \end{tabular}

    \caption{Ablation study on \imaginer adapters.}
    \label{tab:adapter}
\end{table} 

\section{Conclusion}
In this work, we present \ourmethod, the pioneer to enhance the abductive reasoning capabilities of \mllms from verbal and pictorial perspectives. 
We propose a causality-aware contrastive learning algorithm to mine hypotheses with high causal relevance, reducing reasoning space and providing verbal priors for \mllms (\reasoner). Unlike prior methods that rely solely on verbal abduction, we incorporate pictorial thinking via an adapted diffusion model (\imaginer). By jointly training the two components, our method effectively emulates the human-like interplay between language and imagination. 
Extensive experiments on standard benchmarks show that our method consistently outperforms existing small-scale models and competitive MLLM baselines, setting the new state-of-the-art.


\appendix

\bibliography{aaai2026}

@STRING{ACMMM="ACM MM"}

@STRING{CVPR="CVPR"}

@STRING{ECCV="ECCV"}

@STRING{ICCV="ICCV"}

@STRING{WACV="WACV"}

@STRING{ICML="ICML"}

@STRING{ICLR="ICLR"}

@STRING{NIPS="NeurIPS"}

@STRING{ACL="ACL"}

@STRING{AAAI="AAAI"}

@STRING{TIP="IEEE Trans. Image Process."}

@STRING{ACLW = "ACL Workshop"}

@STRING{EMNLP = "Empirical Methods in Natural Language Processing"}

@inproceedings{wu2023multimodal,
	title={Multimodal large language models: A survey},
	author={Wu, Jiayang and Gan, Wensheng and Chen, Zefeng and Wan, Shicheng and Yu, Philip S},
	booktitle={IEEE BigData},
	year={2023}
}

@article{chinchure2024black,
	title={Black Swan: Abductive and Defeasible Video Reasoning in Unpredictable Events},
	author={Chinchure, Aditya and Ravi, Sahithya and Ng, Raymond and Shwartz, Vered and Li, Boyang and Sigal, Leonid},
	journal={arXiv preprint arXiv:2412.05725},
	year={2024}
}

@article{wang2024exploring,
	title={Exploring the reasoning abilities of multimodal large language models (mllms): A comprehensive survey on emerging trends in multimodal reasoning},
	author={Wang, Yiqi and Chen, Wentao and Han, Xiaotian and Lin, Xudong and Zhao, Haiteng and Liu, Yongfei and Zhai, Bohan and Yuan, Jianbo and You, Quanzeng and Yang, Hongxia},
	journal={arXiv preprint arXiv:2401.06805},
	year={2024}
}

@article{tan2025kn,
	title={KN-VLM: KNowledge-guided Vision-and-Language Model for visual abductive reasoning},
	author={Tan, Kuo and Qi, Zhaobo and Zhong, Jianping and Xu, Yuanrong and Zhang, Weigang},
	journal={Multimedia Systems},
	volume={31},
	number={2},
	pages={146},
	year={2025}
}

@article{zhao2022videoabc,
	title={Videoabc: A real-world video dataset for abductive visual reasoning},
	author={Zhao, Wenliang and Rao, Yongming and Tang, Yansong and Zhou, Jie and Lu, Jiwen},
	journal=TIP,
	volume={31},
	pages={6048--6061},
	year={2022}
}

@article{xu2023active,
	title={Active reasoning in an open-world environment},
	author={Xu, Manjie and Jiang, Guangyuan and Liang, Wei and Zhang, Chi and Zhu, Yixin},
	journal=NIPS,
	year={2023}
}

@inproceedings{xu2024probabilistic,
	title={Probabilistic Distillation Transformer: Modelling Uncertainties for Visual Abductive Reasoning},
	author={Xu, Wanru and Miao, Zhenjiang and Tian, Yi and Cen, Yigang and Wan, Lili and Xiaole, Ma},
	booktitle=ACMMM,
	year={2024}
}

@inproceedings{tan2025inferring,
	title={Inferring Past Human Actions in Homes with Abductive Reasoning},
	author={Tan, Clement and Yeo, Chai Kiat and Tan, Cheston and Fernando, Basura},
	booktitle=WACV,
	year={2025},
}

@inproceedings{zhang2024rca,
	title={RCA: Region Conditioned Adaptation for Visual Abductive Reasoning},
	author={Zhang, Hao and Ee, Yeo Keat and Fernando, Basura},
	booktitle=ACMMM,
	year={2024}
}

@inproceedings{li2023multi,
	title={Multi-modal action chain abductive reasoning},
	author={Li, Mengze and Wang, Tianbao and Xu, Jiahe and Han, Kairong and Zhang, Shengyu and Zhao, Zhou and Miao, Jiaxu and Zhang, Wenqiao and Pu, Shiliang and Wu, Fei},
	booktitle=ACL,
	year={2023}
}

@inproceedings{hessel2022abduction,
	title={The abduction of sherlock holmes: A dataset for visual abductive reasoning},
	author={Hessel, Jack and Hwang, Jena D and Park, Jae Sung and Zellers, Rowan and Bhagavatula, Chandra and Rohrbach, Anna and Saenko, Kate and Choi, Yejin},
	booktitle=ECCV,
	year={2022}
}

@article{shanahan2005perception,
	title={Perception as abduction: Turning sensor data into meaningful representation},
	author={Shanahan, Murray},
	journal={Cognitive Science},
	volume={29},
	number={1},
	pages={103--134},
	year={2005}
}

@book{peirce1935collected,
	title={Collected papers of charles sanders peirce},
	author={Peirce, Charles Sanders},
	year={1931},
	publisher={Harvard University Press}
}

@inproceedings{thagard1997abductive,
	title={Abductive reasoning: Logic, visual thinking, and coherence},
	author={Thagard, Paul and Shelley, Cameron},
	booktitle={CLMPST},
	pages={413--427},
	year={1997}
}

@inproceedings{liang2022visual,
  title={Visual abductive reasoning},
  author={Liang, Chen and Wang, Wenguan and Zhou, Tianfei and Yang, Yi},
  booktitle=CVPR,
  year={2022}
}

@article{wang2024qwen2,
  title={Qwen2-vl: Enhancing vision-language model's perception of the world at any resolution},
  author={Wang, Peng and Bai, Shuai and Tan, Sinan and Wang, Shijie and Fan, Zhihao and Bai, Jinze and Chen, Keqin and Liu, Xuejing and Wang, Jialin and Ge, Wenbin and others},
  journal={arXiv preprint arXiv:2409.12191},
  year={2024}
}

@article{zhang2019bertscore,
  title={Bertscore: Evaluating text generation with bert},
  author={Zhang, Tianyi and Kishore, Varsha and Wu, Felix and Weinberger, Kilian Q and Artzi, Yoav},
  journal={arXiv preprint arXiv:1904.09675},
  year={2019}
}

@inproceedings{rombach2022high,
  title={High-resolution image synthesis with latent diffusion models},
  author={Rombach, Robin and Blattmann, Andreas and Lorenz, Dominik and Esser, Patrick and Ommer, Bj{\"o}rn},
  booktitle=CVPR,
  year={2022}
}

@inproceedings{xing2024simda,
  title={Simda: Simple diffusion adapter for efficient video generation},
  author={Xing, Zhen and Dai, Qi and Hu, Han and Wu, Zuxuan and Jiang, Yu-Gang},
  booktitle=CVPR,
  year={2024}
}

@inproceedings{zhou2018end,
  title={End-to-end dense video captioning with masked transformer},
  author={Zhou, Luowei and Zhou, Yingbo and Corso, Jason J and Socher, Richard and Xiong, Caiming},
  booktitle=CVPR,
  year={2018}
}

@inproceedings{xiong2018move,
  title={Move forward and tell: A progressive generator of video descriptions},
  author={Xiong, Yilei and Dai, Bo and Lin, Dahua},
  booktitle=ECCV,
  year={2018}
}

@inproceedings{dai2019transformer,
  title={Transformer-xl: Attentive language models beyond a fixed-length context},
  author={Dai, Zihang and Yang, Zhilin and Yang, Yiming and Carbonell, Jaime and Le, Quoc V and Salakhutdinov, Ruslan},
  booktitle=ACL,
  year={2019}
}

@inproceedings{lei2020mart,
  title={Mart: Memory-augmented recurrent transformer for coherent video paragraph captioning},
  author={Lei, Jie and Wang, Liwei and Shen, Yelong and Yu, Dong and Berg, Tamara L and Bansal, Mohit},
  booktitle=ACL,
  year={2020}
}

@inproceedings{wang2021end,
  title={End-to-end dense video captioning with parallel decoding},
  author={Wang, Teng and Zhang, Ruimao and Lu, Zhichao and Zheng, Feng and Cheng, Ran and Luo, Ping},
  booktitle=ICCV,
  year={2021}
}

@inproceedings{wu2024number,
  title={Number it: Temporal Grounding Videos like Flipping Manga},
  author={Wu, Yongliang and Hu, Xinting and Sun, Yuyang and Zhou, Yizhou and Zhu, Wenbo and Rao, Fengyun and Schiele, Bernt and Yang, Xu},
  booktitle=CVPR,
  year={2025}
}

@inproceedings{hang2023efficient,
  title={Efficient diffusion training via min-snr weighting strategy},
  author={Hang, Tiankai and Gu, Shuyang and Li, Chen and Bao, Jianmin and Chen, Dong and Hu, Han and Geng, Xin and Guo, Baining},
  booktitle=ICCV,
  year={2023}
}

@inproceedings{chen2020simple,
  title={A simple framework for contrastive learning of visual representations},
  author={Chen, Ting and Kornblith, Simon and Norouzi, Mohammad and Hinton, Geoffrey},
  booktitle=ICML,
  year={2020}
}

@inproceedings{li2024cross,
  title={Cross-modal Observation Hypothesis Inference},
  author={Li, Mengze and Han, Kairong and Xu, Jiahe and Li, Yueying and Wu, Tao and Zhao, Zhou and Miao, Jiaxu and Zhang, Shengyu and Chen, Jingyuan},
  booktitle=ACMMM,
  year={2024}
}

@inproceedings{wang2024internvideo2,
	title={Internvideo2: Scaling video foundation models for multimodal video understanding},
	author={Wang, Yi and Li, Kunchang and Li, Xinhao and Yu, Jiashuo and He, Yinan and Wang, Chenting and Chen, Guo and Pei, Baoqi and Zheng, Rongkun and Xu, Jilan and Wang, Zun and others},
	booktitle=ECCV,
	year={2024}
}

@inproceedings{lin2024video,
	title={Video-LLaVA: Learning United Visual Representation by Alignment Before Projection},
	author={Lin, Bin and Ye, Yang and Zhu, Bin and Cui, Jiaxi and Ning, Munan and Jin, Peng and Yuan, Li},
	booktitle=EMNLP,
	year={2024}
}

@article{lyu2023macaw,
	title={Macaw-LLM: Multi-Modal Language Modeling with Image, Audio, Video, and Text Integration},
	author={Lyu, Chenyang and Wu, Minghao and Wang, Longyue and Huang, Xinting and Liu, Bingshuai and Du, Zefeng and Shi, Shuming and Tu, Zhaopeng},
	journal={arXiv preprint arXiv:2306.09093},
	year={2023}
}

@article{chen2023x,
	title={X-LLM: Bootstrapping Advanced Large Language Models by Treating Multi-Modalities as Foreign Languages},
	author={Chen, Feilong and Han, Minglun and Zhao, Haozhi and Zhang, Qingyang and Shi, Jing and Xu, Shuang and Xu, Bo},
	journal={arXiv preprint arXiv:2305.04160},
	year={2023}
}

@article{zohar2024apollo,
	title={Apollo: An Exploration of Video Understanding in Large Multimodal Models},
	author={Orr Zohar and Xiaohan Wang and Yann Dubois and Nikhil Mehta and Tong Xiao and Philippe Hansen-Estruch and Licheng Yu and Xiaofang Wang and Felix Juefei-Xu and Ning Zhang and Serena Yeung-Levy and Xide Xia},
	journal={arXiv preprint arXiv:2412.10360},
	year={2024}
}

@inproceedings{Maaz2024VideoChatGPT,
	title={Video-ChatGPT: Towards Detailed Video Understanding via Large Vision and Language Models},
	author={Maaz, Muhammad and Rasheed, Hanoona and Khan, Salman and Khan, Fahad Shahbaz},
	booktitle=ACL,
	year={2024}
}

@inproceedings{Wang2024VideoCoT,
	title={VideoCoT: A Video Chain-of-Thought Dataset with Active Annotation Tool},
	author={Yan Wang and Yawen Zeng and Jingsheng Zheng and Xiaofen Xing and Jin Xu and Xiangmin Xu},
	booktitle=ACLW,
	year={2024}
}

@article{damonlpsg2024videollama2,
	title={VideoLLaMA 2: Advancing Spatial-Temporal Modeling and Audio Understanding in Video-LLMs},
	author={Cheng, Zesen and Leng, Sicong and Zhang, Hang and Xin, Yifei and Li, Xin and Chen, Guanzheng and Zhu, Yongxin and Zhang, Wenqi and Luo, Ziyang and Zhao, Deli and Bing, Lidong},
	journal={arXiv preprint arXiv:2406.07476},
	year={2024}
}

@article{Xu2023YoukumPLUGA1,
	title={Youku-mPLUG: A 10 Million Large-scale Chinese Video-Language Dataset for Pre-training and Benchmarks},
	author={Haiyang Xu and Qinghao Ye and Xuan-Wei Wu and Mingshi Yan and Yuan Miao and Jiabo Ye and Guohai Xu and Anwen Hu and Yaya Shi and Guangwei Xu and Chenliang Li and Qingfang Qian and Maofei Que and Ji Zhang and Xiaoyan Zeng and Feiyan Huang},
	journal={arXiv preprint arXiv:2306.04362},
	year={2023}
}

@article{Luo2023ValleyVA,
	title={Valley: Video Assistant with Large Language model Enhanced abilitY},
	author={Ruipu Luo and Ziwang Zhao and Min Yang and Junwei Dong and Ming-Hui Qiu and Pengcheng Lu and Tao Wang and Zhongyu Wei},
	journal={arXiv preprint arXiv:2306.07207},
	year={2023}
}

@inproceedings{li2024mvbench,
  title={Mvbench: A comprehensive multi-modal video understanding benchmark},
  author={Li, Kunchang and Wang, Yali and He, Yinan and Li, Yizhuo and Wang, Yi and Liu, Yi and Wang, Zun and Xu, Jilan and Chen, Guo and Luo, Ping and others},
  booktitle=CVPR,
  year={2024}
}

@inproceedings{zhou2018towards,
  title={Towards automatic learning of procedures from web instructional videos},
  author={Zhou, Luowei and Xu, Chenliang and Corso, Jason},
  booktitle=AAAI,
  year={2018}
}

@article{singer2022make,
  title={Make-a-video: Text-to-video generation without text-video data},
  author={Singer, Uriel and Polyak, Adam and Hayes, Thomas and Yin, Xi and An, Jie and Zhang, Songyang and Hu, Qiyuan and Yang, Harry and Ashual, Oron and Gafni, Oran and others},
  journal=ICLR,
  year={2022}
}

@inproceedings{blattmann2023align,
  title={Align your latents: High-resolution video synthesis with latent diffusion models},
  author={Blattmann, Andreas and Rombach, Robin and Ling, Huan and Dockhorn, Tim and Kim, Seung Wook and Fidler, Sanja and Kreis, Karsten},
  booktitle=CVPR,
  year={2023}
}

@inproceedings{he2016deep,
  title={Deep residual learning for image recognition},
  author={He, Kaiming and Zhang, Xiangyu and Ren, Shaoqing and Sun, Jian},
  booktitle=CVPR,
  year={2016}
}

@inproceedings{ioffe2015batch,
  title={Batch normalization: Accelerating deep network training by reducing internal covariate shift},
  author={Ioffe, Sergey and Szegedy, Christian},
  booktitle=ICML,
  year={2015},
}

@article{wang2024diffusion,
  title={Diffusion feedback helps clip see better},
  author={Wang, Wenxuan and Sun, Quan and Zhang, Fan and Tang, Yepeng and Liu, Jing and Wang, Xinlong},
  journal=ICLR,
  year={2025}
}

@article{ma2025genhancer,
  title={GenHancer: Imperfect Generative Models are Secretly Strong Vision-Centric Enhancers},
  author={Ma, Shijie and Ge, Yuying and Wang, Teng and Guo, Yuxin and Ge, Yixiao and Shan, Ying},
  journal={CoRR},
  year={2025}
}

@article{hurst2024gpt,
  title={Gpt-4o system card},
  author={Hurst, Aaron and Lerer, Adam and Goucher, Adam P and Perelman, Adam and Ramesh, Aditya and Clark, Aidan and Ostrow, AJ and Welihinda, Akila and Hayes, Alan and Radford, Alec and others},
  journal={arXiv preprint arXiv:2410.21276},
  year={2024}
}

@inproceedings{robin2021contrast,
  title={CONTRASTIVE LEARNING WITH HARD NEGATIVE SAMPLES},
  author={Robinson, Joshua and Chuang, Ching-Yao and Sra, Suvrit and Jegelka, Stefanie},
  booktitle=ICLR,
  year={2021}
}

@article{wang2025videorft,
  title={VideoRFT: Incentivizing Video Reasoning Capability in MLLMs via Reinforced Fine-Tuning},
  author={Wang, Qi and Yu, Yanrui and Yuan, Ye and Mao, Rui and Zhou, Tianfei},
  journal=NIPS,
  year={2025}
}

\end{document}